\documentclass[runningheads]{llncs}
\usepackage{graphicx}
\graphicspath{{./figure/}}
\usepackage{caption}
\usepackage{float}
\usepackage{booktabs}
\usepackage{makecell}
\usepackage{subcaption}
\usepackage{amsmath}
\usepackage[table]{xcolor}
\usepackage{tikz}
\usetikzlibrary{trees,positioning}
\usepackage{nicematrix}
\usepackage{multirow}
\usepackage{pifont}
\usepackage[colorlinks=true,linkcolor=blue,citecolor=blue]{hyperref}
\usepackage[bottom]{footmisc}
\definecolor{darkgrey}{HTML}{898989}
\definecolor{lightblue}{HTML}{bbdefb}
\definecolor{lightgrey}{HTML}{efefef}
\definecolor{lightlightblue}{HTML}{d2edfd}

\begin{document}
\title{A Survey on Table Question Answering: Recent Advances}

\author{Nengzheng Jin\inst{1} \and Joanna Siebert\inst{1} \and Dongfang Li\inst{1} \and Qingcai Chen\thanks{Corresponding author.}\inst{1,2}}
\authorrunning{N. Jin et al.}
\institute{Harbin Institute of Technology (Shenzhen), Shenzhen, China \and Peng Cheng Laboratory, Shenzhen, China \\
\email{\{nengzhengjin,joannasiebert,crazyofapple\}@gmail.com, qingcai.chen@hit.edu.cn}}
\maketitle
\begin{abstract}
Table Question Answering (Table QA) refers to providing precise answers from tables to answer a user's question. In recent years, there have been a lot of works on table QA, but there is a lack of comprehensive surveys on this research topic. Hence, we aim to provide an overview of available datasets and representative methods in table QA. We classify existing methods for table QA into five categories according to their techniques, which include semantic-parsing-based, generative, extractive, matching-based, and retriever-reader-based methods. Moreover, because table QA is still a challenging task for existing methods, we also identify and outline several key challenges and discuss the potential future directions of table QA.

\end{abstract}

\section{Introduction}
\begin{figure}[b]
    \centering
    \footnotesize
    \begin{minipage}[t]{\textwidth}
    \centering
    \colorbox{lightgrey}{\textbf{Question:} What was the reported mainline RPM for American Airlines in 2017?}
    \\ \  \\
    \end{minipage}
    \begin{minipage}[b]{\textwidth}
    \centering
    \setlength{\tabcolsep}{0.7em} % for the horizontal padding
    \begin{NiceTabular}{lrrr}
     Table 1.&\multicolumn{3}{c}{Year Ended December 31.} \\
     \cline{2-4}
     &2017&2016&2015\\
     \cline{2-4}
     \rowcolor{lightlightblue}
     \textbf{Mainline}&&& \\
     \ \ Revenue passenger miles (millions)$^{(a)}$&\textcolor{blue!50}{\textbf{201,351}}&199,014&199,467 \\
     \rowcolor{lightlightblue}
     \ \ Available seat miles (millions)$^{(b)}$&243,806&241,734&239,375 \\
     \ \ Passenger load factor (percent)$^{(c)}$&82.6&82.3&83.3 \\
     \hline
    \end{NiceTabular}
    \end{minipage}
    \caption{An illustration example of table QA (tailored from \cite{katsisAITQAQuestionAnswering2021}). The bold number (201,351) is the target answer.}
    \label{fig:illustration_example}
\end{figure}
\textbf{Tables}, which are an effective way to store and present data, are pervasive in various real-world scenarios, for example, financial reports and scientific papers. To leverage valuable information in tables, recent studies have applied \textbf{table question answering} as one important technique \cite{muellerAnsweringConversationalQuestions2019,pasupat-liang-2015-compositional,Zhu2021TATQAAQ}. Given the user's question, table QA aims to provide precise answers through table understanding and reasoning. For example, Figure \ref{fig:illustration_example} illustrates the question answering over the tables from airline industry.

Generally speaking, table QA tasks can be traced back to querying relational databases with natural language, in which the tables are relatively structured. In this case, table QA task is solved by using a semantic parser that transforms natural language into a structured query (e.g., SQL), then executing it to retrieve answers\cite{Dong2016LanguageTL,Zhong2017Seq2SQLGS}. For the tables that do not come from a database (non-database tables, e.g., web tables, spreadsheet tables), researchers also treat semantic parsing as an important method \cite{Liang2018MemoryAP,pasupat-liang-2015-compositional}. However, for tables with surrounding text,  some methods directly extract \cite{eisenschlosMATEMultiviewAttention2021,herzigTaPasWeaklySupervised2020,Zhu2021TATQAAQ} or generate answers \cite{chenFinQADatasetNumerical2021a,nanFeTaQAFreeformTable2021} from the tables and their surrounding text without generating a structured query. For open-domain table QA, a retriever is needed to retrieve the related tables from a large corpus; then a reader is used to produce the answers from the retrieved tables \cite{chenOpenQuestionAnswering2021,herzigOpenDomainQuestion2021}.

Despite plenty of studies being conducted on table QA, there is a lack of systematic survey of this research field. In contrast, for entirely structured data (e.g., knowledge base), \cite{Fu2020ASO} provides a detailed survey of knowledge base question answering (KBQA). For unstructured text, \cite{abbasiantaeb2021text} discusses the existing approaches and challenges of text-based question answering. Hence, our survey aims to provide a relatively thorough introduction to related datasets, existing methods and challenges of table QA, to help researchers grasp the recent advancements.

The main contributions of our survey are as follows: (1) we present most of the available datasets of table QA and create a mapping between each dataset and existing methods, to show which methods can be applied on a given dataset. (2) We summarize five kinds of table QA methods and make a relatively thorough introduction of each one. (3) We identify and discuss two main challenges of table encoding and table reasoning, which might be helpful for future research.

The remainder of the paper is organized as follows: In Section \ref{section:background}, we introduce the preliminary knowledge of table QA. In Section \ref{section:datasets}, we provide an overview of the available datasets. After that, we introduce semantic-parsing-based methods individually in Section \ref{section:semantic parsing} because they are a big group, and introduce the rest of methods in Section \ref{section:non semantic parsing}. Finally, in Section \ref{section:challengs}, we discuss several key challenges and potential future directions to explore.

\section{Background}
\label{section:background}
This section provides preliminary information for an in-depth understanding of table QA, including the composition of tables, definitions of semantic parsing and table question answering.

\subsection{The Composition of Tables}

A table can be seen as a grid of cells arranged in rows and columns \cite{Zhang2020WebTE}. Tables that come from database are relatively structured and consist of several columns (i.e., attributes). However, there are additional elements in web tables \cite{Zhang2020WebTE}, including page title, caption, headers, and so forth. Considering that tabular data is usually surrounded by textual annotations in table QA, we refer two additional elements called pre-annotation and post-annotation as supplement parts of a table. Specifically, pre/post-annotation refers to the related sentences that appear before or after a table.

\subsection{Semantic Parsing}
Semantic parsing refers to transforming the natural language utterance into a logical form that can be executed by machines. One of classical semantic parsing tasks is text2sql, which converts the natural language utterances into structured query language (SQL). For instance, the question ``What’s the lowest pick in round 1?" should be transformed into ``SELECT MIN(Pick) FROM mytable WHERE Rnd = 1;".

\subsection{Table Question Answering}

In this subsection, we disentangle the relationship between KBQA, text-based QA and table QA, and introduce two classifications of table QA.

Three QA tasks aim to provide answers to a user's question. The main difference is their reference source. KBQA is conducted over knowledge base, which is regarded as a kind of structured knowledge, text-based QA is conducted over unstructured text, and table QA is conducted over non-database tables which are regarded as semi-structured knowledge \cite{pasupat-liang-2015-compositional}, as well as over database tables which are relatively structured. We introduce two classifications of table QA below.

\textbf{Open-domain vs. Closed-domain}. This classification is for web tables. Open-domain table QA answers the question based on large-scale table documents.  In contrast, closed-domain table QA answers the question based on a limited number of tables (usually one table).

\textbf{Free-form vs. Non-free-form}. This classification is based on the form of the answers. Give the user's question, free-form table QA requires generating dialog-like answers from the tables  \cite{nanFeTaQAFreeformTable2021}. However, non-free-form table QA aims to provide a factual answer, which usually consists of a few words. 

\section{Overview of Datasets}
\label{section:datasets}

The research on table QA has been increasing over the past few years due to the availability of large datasets. In this survey, we present an overview of these table QA datasets. As shown in Table \ref{tab:datasets}, most of the datasets are closed-domain, and their question type is factoid \footnotemark[1]\footnotetext[1]{We extend the question type with free form and multiple choice, which originally includes factoid, list, definition and complex.}. 

\begin{table}[htb]
    \centering
    \footnotesize
    \setlength{\tabcolsep}{3pt}
    \renewcommand{\arraystretch}{1.2}
    \caption{An overview of table QA datasets. The representative methods without marks (e.g. \ding{61}\ding{75}\ding{162}) can be used on the datasets aligned in the same horizontal zone, and the methods with marks are currently adopted on the datasets with the same mark.}
    \begin{tabular}{|l|l|c|l|l|}
        \hline
        \multicolumn{2}{|c|}{\textbf{Dataset}}&\textbf{\makecell[c]{Closed\\-domain}}&\textbf{\makecell[c]{Question\\ Type}}& \makecell[l]{\textbf{Representative} \textbf{Methods}} \\
        \hline
        \multirow{6}{*}{\textbf{\makecell[c]{Table\\-only}}} &  WTQ\ding{75} \cite{pasupat-liang-2015-compositional} & Yes & Factoid & \multirow{6}{12em}{Semantic parsing-based \cite{Dasigi2019IterativeSF,Dong2016LanguageTL,Dong2018CoarsetoFineDF,He2019XSQLRS,Hwang2019ACE,iyyerSearchbasedNeuralStructured2017,Liang2018MemoryAP,neelakantanNeuralProgrammerInducing2016,pasupat-liang-2015-compositional}\\
        \cite{Sun2018SemanticPW,Wang2018RobustTG,Xu2017SQLNetGS,yinNeuralEnquirerLearning2016,Yu2018TypeSQLKT,Zhong2017Seq2SQLGS}\\
        Generative method\ding{162} \cite{muellerAnsweringConversationalQuestions2019}
        Matching-based method\ding{61} \cite{glassCapturingRowColumn2021}\\
        Extractive method\ding{75} \cite{herzigTaPasWeaklySupervised2020}      }\\
         &SQA\ding{162}\ding{75} \cite{iyyerSearchbasedNeuralStructured2017}& Yes & Factoid & \\
         &WikiSQL\ding{75} \cite{Zhong2017Seq2SQLGS} & Yes & Factoid & \\
         &Spider \cite{Yu2018SpiderAL}&Yes&Factoid& \\
        %  \cline{5-5}
         &HiTab \cite{Cheng2021HiTabAH} & Yes&Factoid& \\
         &AIT-QA\ding{61}\ding{75} \cite{katsisAITQAQuestionAnswering2021}&Yes&Factoid& \\
        \hline
        \multirow{7}{*}{\textbf{\makecell[c]{Non\\-table\\-only}}} & FeTaQA\cite{nanFeTaQAFreeformTable2021}&Yes&Free form&Generative method \cite{nanFeTaQAFreeformTable2021} \\
        \cline{2-5}
		 &FinQA \cite{chenFinQADatasetNumerical2021a}&Yes&Factoid&Semantic parsing-based\cite{chenFinQADatasetNumerical2021a}\\
		 \cline{2-5}
		 &TAT-QA \cite{Zhu2021TATQAAQ}&Yes&Factoid&\multirow{2}{12em}{Extractive methods \cite{chenHybridQADatasetMultiHop2020,eisenschlosMATEMultiviewAttention2021,Zhu2021TATQAAQ}}\\
		 &HybridQA \cite{chenHybridQADatasetMultiHop2020}&Yes&Factoid&\\
		 \cline{2-5}
         &TabMCQ \cite{jauharTabMCQDatasetGeneral2016}&Yes&Multiple choice&\multirow{2}{12em}{Matching-based methods \cite{jauharTabMCQDatasetGeneral2016,Li2021TSQATS}}\\
         &GeoTSQA \cite{Li2021TSQATS}&Yes&Multiple choice&\\
        \cline{2-5}
         &OTTQA \cite{chenOpenQuestionAnswering2021}&No&Factoid&\multirow{2}{12em}{Retriever-reader-based methods \cite{chenOpenQuestionAnswering2021,herzigOpenDomainQuestion2021,Li2021DualRO,oguzUniKQAUnifiedRepresentations2021,zhongReasoningHybridChain2022}}\\
         &NQ-tables \cite{herzigOpenDomainQuestion2021}&No&Factoid&\\
        \hline
    \end{tabular}
    \vspace{5pt}
    
    \label{tab:datasets}
\end{table}

\textbf{Table-only Datasets} contain database tables or non-database tables without pre/post-annotation. These datasets contain different kinds of supervision for model training. Some datasets, such as WikiSQL \cite{Zhong2017Seq2SQLGS} and Spider\cite{Yu2018SpiderAL}, provide logical form annotations as supervision. However, others provide the final answers as supervision, for example, WTQ \cite{pasupat-liang-2015-compositional} and SQA \cite{Cheng2021HiTabAH}.
Most of the table-only datasets consist of relational tables with regular structure except HiTab \cite{Cheng2021HiTabAH} and AITQA \cite{katsisAITQAQuestionAnswering2021}, whose tables have hierarchical structure and a number of merged cells.  

\textbf{Non-table-only Datasets} include samples that consist of a table and its pre/post-annotation. Among these datasets, OTTQA \cite{chenOpenQuestionAnswering2021} and NQ-tables  \cite{herzigOpenDomainQuestion2021}, which are used for open-domain table QA, are constructed from existing closed-domain datasets. TAT-QA \cite{Zhu2021TATQAAQ} and FinQA \cite{chenFinQADatasetNumerical2021a} are extracted from financial reports with a large number of tables. Moreover, researchers propose TabMCQ \cite{jauharTabMCQDatasetGeneral2016} and GeoTSQA \cite{Li2021TSQATS} datasets that contain multiple choice questions. Generally, non-table-only datasets require modelling over tables and text, which have been a popular and challenging research topic in recent years.

\section{Semantic-parsing-based Methods}
\label{section:semantic parsing}

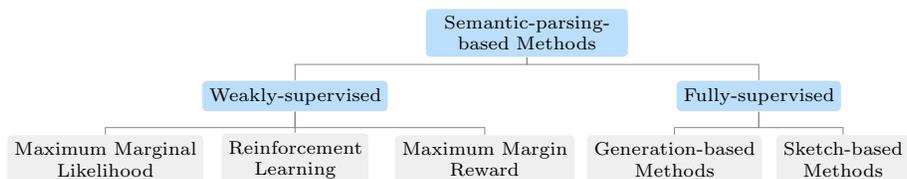
\begin{figure}
    \centering
    \scriptsize
    \begin{tikzpicture}[
    edge from parent fork down,align=center,font=\scriptsize,level distance=3em,
    every node/.style={rounded corners=0.6ex},
    edge from parent/.style={draw=darkgrey,ultra thin},
    level 1/.style={sibling distance=22em}]
    \definecolor{lightblue}{HTML}{bbdefb}
    \definecolor{lightgrey}{HTML}{efefef}
    \node[fill=lightblue,text width=2.5cm]{Semantic-parsing-\\based Methods}
        child{node[fill=lightblue]{Weakly-supervised}
            child[sibling distance=9em]{node[fill=lightgrey]{Maximum Marginal\\Likelihood}}
            child[sibling distance=9em]{node[fill=lightgrey]{Reinforcement\\Learning}}
            child[sibling distance=9em]{node[fill=lightgrey]{Maximum Margin\\Reward}}
            }
        child{node[fill=lightblue]{Fully-supervised}
            child[sibling distance=8em]{node[fill=lightgrey]{Generation-based\\Methods}}
            child[sibling distance=8em]{node[fill=lightgrey]{Sketch-based\\Methods}}
            };
    \end{tikzpicture}
    \caption{The overview of semantic-parsing-based methods for table QA.}
    \label{fig:semantic-parsing}
\end{figure}

In table QA tasks, the semantic-parsing-based methods first transform the question into a logical form (e.g., SQL), and then execute the logical form on tables to retrieve the final answer. These methods can be categorized into weakly-supervised and fully-supervised methods. In the weakly supervised setting, given the question $q$ and table $t$, the semantic parser for table QA is to generate the logical form $y$ with weak supervision of the final answer $z$. In this setting, no gold logical forms are provided. However, in the fully-supervised setting, logical forms that execute to the correct answers will be provided as stronger supervision. We depict two of these methods in Figure \ref{fig:semantic-parsing} and discuss them in more detail below.

\subsection{Weakly-supervised Table Semantic Parsing}
In earlier research, weakly-supervised semantic parsing for table QA was based on hand-crafted features and \textbf{grammar rules}. For instance, Floating Parser \cite{pasupat-liang-2015-compositional} builds logical forms by applying predefined deduction rules recursively; it uses beam search to contain a limited number of partial logical forms at each recursion and prunes invalid partial logical forms with type constraints. The final logical form will be executed on related tables to obtain the answers.

However, the Floating Parser is domain-specific and only a few of learnable parameters of it can be adjusted. Hence, Yin et al. \cite{yinNeuralEnquirerLearning2016} propose a more powerful semantic parser, Neural Enquirer. They use a query encoder and table encoder to encode the question and table, respectively. Then, an executor implemented with DNN generates the partial logical form, and final result will be generated by the final layer. Similarly, Neelakantan et al. \cite{neelakantanLearningNaturalLanguage2017} provide an approach based on Neural Programmer \cite{neelakantanNeuralProgrammerInducing2016}. Neural Programmer takes the hidden states of question RNN and history RNN as input and selects an operation and column that are related to the final answers. In practice, \cite{neelakantanLearningNaturalLanguage2017,neelakantanNeuralProgrammerInducing2016} find it difficult to train the model under weak supervision. Hence, some learning strategies \cite{iyyerSearchbasedNeuralStructured2017,krishnamurthyNeuralSemanticParsing2017,Zhong2017Seq2SQLGS} are proposed to tackle the problem, which can be categorized into \textit{maximum marginal likelihood}, \textit{reinforcement learning}, and \textit{maximum margin reward}.

\textbf{Maximum Marginal Likelihood (MML)}: the MML strategy in a weakly-supervised setting is to optimize the objective function $\log p(z_i|x_i,t_i)$ , which represents the probability of generating the correct answer $z_i$ by using candidate logical forms $Y$, based on question $x_i$ and table $t_i$.

\begin{equation}\label{mml_objective}
    J_{M M L}=\log p\left(z_{i} \mid x_{i}, t_i\right)=\log \sum_{y \in Y} p\left(z_{i} \mid y, t_{i}\right) p\left(y \mid x_{i}, t_{i}\right)
\end{equation}

The MML algorithm is usually adapted with an \textbf{encoder–decoder neural network}, where the encoder encodes the questions and tables, and the decoder generates the logical forms step by step \cite{krishnamurthyNeuralSemanticParsing2017,Wang2019LearningSP}. Based on this, Dasigi et al. \cite{Dasigi2019IterativeSF} extends MML training with a coverage-augmented loss function. The usage of coverage-augmented training and searching provides extra guidance to search for more consistent logical forms under binary supervision signals, which leads to better performance. Recently, Min et al. \cite{Min2019ADH} have proposed a variation of MML (called HardEM) that simplifies the sum operation in the MML objectvie with a max operation and that outperforms previous methods in several QA tasks. 

\textbf{Reinforcement Learning (RL)}: In this learning strategy, semantic parsing for table QA is accomplished by an agent that takes a sequence of actions based on a certain policy. In practice, the policy is initialized as stochastic and trained with the goal of maximizing the expected reward. The expected reward for the sample $(x_i,z_i,t_i)$ is shown in Equation \ref{equation:RL}.
\begin{equation}\label{equation:RL}
    J_{R L}=\sum_{y \in Y} p\left(y \mid x_{i}, t_{i}\right) R\left(y, z_{i}\right)
\end{equation}

Zhong et al. \cite{Zhong2017Seq2SQLGS} are the first to apply policy gradient, an important method of reinforcement learning, on the WikiSQL-weak dataset. It adopts the attentional sequence to sequence neural semantic parser \cite{Dong2016LanguageTL} and augment this model with a pointer network to enable copying the input symbol as a part of the output. To further improve the performance, Misra et al. \cite{Misra2018PolicySA} adopt policy shaping that refers to introducing prior knowledge into a policy. They propose two kinds of prior knowledge—$match(x,y)$ and $co\_occur(y,x)$, which help to select the logical form with the higher probability of being correct. Additionally, the semantic parser MAPO\cite{Liang2018MemoryAP} utilizes memory buffer to store the logical forms (trajectories) with high rewards for stabilizing and accelerating the model training. 

\textbf{Maximum Margin Reward (MMR)}: The MMR strategy maximizes a margin objective, which only update the score of the highest scoring logical form and the logical form that violates the margin constraint the most. Iyyer at el. \cite{iyyerSearchbasedNeuralStructured2017} use this learning strategy in DynSP model \cite{iyyerSearchbasedNeuralStructured2017} , which achieves better performance against previous methods on SQA dataset.

\subsection{Fully-Supervised Table Semantic Parsing}
\label{subsection:fully-supervised semantic parsing}

We have discussd the weakly supervised semantic parsing in the above subsections. However, there are also a number of methods requiring a fully supervised setting to achieve better performance for real-world applications. In the question answering task over database tables (e.g., WikiSQL), fully supervised semantic parsing can be roughly categorized into generation-based methods and sketch-based methods. 

\textbf{Generation-based Methods}:  Generation-based methods usually adopt a sequence-to-sequence (Seq2Seq) framework, where an encoder is used to encode the question and optional table, and the decoder generates the logical form autoregressively \cite{Dong2016LanguageTL,Zhong2017Seq2SQLGS}. Based on this framework, Sun et al. \cite{Sun2018SemanticPW} leverage the structure of a table and syntax of SQL language for better SQL generation. To further constrain the decoder, Yin et al. \cite{yin2017syntactic} utilize a grammar model to evaluate the generation at each timestep and exclude those invalid candidates based on grammar rules. Similarly, Wang et al. \cite{Wang2018RobustTG} detect and exclude faulty programs by conditioning on the execution of partially generated program. Besides, Cho et al. \cite{choAdversarialTableQAAttention2018} replace SQL annotations in WikiSQL dataset with special logical forms that annotate operand information, and improve the robustness of the semantic parser by supervising attention weights through the operand information and using cascade selective unit at each decoding step. Recently, some researchers (e.g., \cite{Cao2021LGESQLLG,Wang2020RATSQLRS}) have proposed schema linking that links the mentions in questions to the schema content; they use relation-aware Transformer or graph neural network (GNN) as the encoder to model the links, which is proved to be an effective method.

\textbf{Sketch-based Methods}: Sketch-based models decompose the target SQL query into several modules. Through performing classification for each module, complete SQL is then recomposed based on the classification results. For example, SQLNet \cite{Xu2017SQLNetGS} formulates the SQL sketch as ``SELECT \$AGG \$COLUMN WHERE \$COLUMN \$OP \$VALUE (AND \$COLUMN \$OP \$VALUE)\*" and predicts the value for each \$ variable through classification. Following this, TypeSQL \cite{Yu2018TypeSQLKT} leverages the type information of the question entities based on external knowledge base to better understand the rare entities. Further, some researchers \cite{He2019XSQLRS,Hwang2019ACE} have adopted pre-trained model as an encoder to better understand the question for the classification, which significantly enhances the performance. In another type of sketch-based method, the Coarse-to-Fine model \cite{Dong2018CoarsetoFineDF} uses two-stage decoding. Unlike SQLNet \cite{Xu2017SQLNetGS}, the sketch in the Coarse-to-Fine model are not statically predefined but generated by an decoder. Based on the sketch, the complete SQL is then generated through the second decoder.  

\section{Non-semantic-parsing-based Methods}
\label{section:non semantic parsing}
\subsection{Generative Methods}
\label{subsection:generative methods}
Notice that there are a number of semantic-parsing-based methods adopting generative models, for example, Seq2Seq neural network \cite{Dong2016LanguageTL,Wang2018RobustTG,Zhong2017Seq2SQLGS}. The main difference between generative methods and Seq2Seq semantic-parsing-based methods is that the former does not generate the logical form, but instead generates the answer directly. Hence, compared with Seq2Seq semantic-parsing-based methods that can handle both database tables and non-database tables, generative methods mainly focus on question answering over non-database tables.

For free-form table QA, a generative model becomes necessary to generate free-form answers. \cite{nanFeTaQAFreeformTable2021} is the first to conduct free-form question answering over tables. It adopts an end-to-end pre-trained model to encode the question and linearized tables, as well as to generate free-form answers. Generative models can also be used for non-free-form table QA. M{\"u}ller et al. \cite{muellerAnsweringConversationalQuestions2019} propose a graph-based generative model for SQA task \cite{iyyerSearchbasedNeuralStructured2017}; they transform tables into graphs by representing the columns, rows and cells as nodes, introducing cell-column and cell-row relation. Then, the graphs are encoded using a graph neural network, and the answers are generated by a Transformer-based decoder. However, the graph-based generative model \cite{muellerAnsweringConversationalQuestions2019} is only feasible for table-only tasks because it dose not model the pre/post-annotation of the tables. Hence, Zayats et al. \cite{zayatsRepresentationsQuestionAnswering2021} extend this method to hybrid question answering tasks. Because it technically belongs to extractive methods, we present this method in the following subsection.

\subsection{Extractive Methods}
\label{subsection:extractive methods}
Rather than generating the answer through a decoder, extractive methods directly select or extract the token spans from the linearized table as candidate answers or evidences. For example, Herzig et al. \cite{herzigTaPasWeaklySupervised2020} use a pre-trained encoder to represent tables and select table cells with the highest probability as answers. Zhu et al. \cite{Zhu2021TATQAAQ} also follow the same paradigm to extract evidences from tables; then, they retrieve answers by simple reasoning over the evidences. 

In this kind of methods, the semantic representations of table cells become important because the model needs to understand which table cells are relevant to the question. To avoid incorporating irrelevant information into the table cell representations, several structure-aware approaches have been proposed by researchers. The simplest way to incorporate structural information into table cell representations is by adding hand-crafted features. For example, TAPAS \cite{herzigTaPasWeaklySupervised2020} uses row/column embedding as an additional input of Transformer to indicate the position of a table cell, which implicitly models the table structure. But this setting does not models the relation between tables cells, which is achieved by the following methods.

\textbf{Attention Mask}: This type of methods model the table structure through attention mask that selectively masking the irrelevant token in self-attention layer of Transformer. Eisenschlos et al. \cite{eisenschlosMATEMultiviewAttention2021} propose a multi-view attention mechanism that splits attention heads into row heads and column heads, where each row/column head only incorporates the information of cells from the same row/column and information from the question into current token. Further, they reorder the input sequence for row heads and column heads separately and apply a windowed attention mechanism. This technique turns quadratic time complexity into a linear one, leading to better structure-aware table representations and smaller time complexity simultaneously.

\textbf{Attention Bias}: Another way to represent the table structure is to inject attention bias into the attention layer of Transformer. For example, Zayats et al. \cite{zayatsRepresentationsQuestionAnswering2021} propose an extractive model based on the Transformer-based GNN model \cite{muellerAnsweringConversationalQuestions2019}. The ``GNN model" is implemented by introducing attention bias when calculating the attention weight. Furthermore, they enrich the table representations with the embedding of the relevant text, leading to more precise and richer table cell representations. Similarly, TableFormer \cite{Yang2022TableFormerRT}, a variation of Transformer, also proposes 13 table-text attention biases into the self-attention layer (e.g., \textit{same row}, \textit{cell to column header}). Through pre-training, this model achieves remarkable performance on several benchmarks.  

\subsection{Matching-Based Methods}
Matching-based models usually process the question and each fragment of the table (e.g., row, cell) individually, and predict the matching score between them. The final answer is retrieved by simple reasoning on the most relevant fragments. For example, Sun et al. \cite{Sun2016TableCS} formulate the table QA task as a joint entity and relation matching problem. They first transform the question and each row of the table into chains, which are two-node graphs. Then, this model matches the question chain to all candidate column chains through snippets matching and deep chain inference, hence retrieving top-$k$ candidate chains for final answer generation. Similarly, in a multiple-choice table QA task, Jauhar et al. \cite{jauharTabMCQDatasetGeneral2016} match each question-choice pair (also referred to QA pair) to the rows of the table and return the highest matching score as the confidence of the QA pair. In a similar way, the RCI model \cite{glassCapturingRowColumn2021} predicts a matching score between a row/column and question. Then, the confidence of a table cell as being the correct answer can be calculated by combining the matching score of its row and column. 

\subsection{Retriever-reader-based Methods}

\begin{figure}[b]
    \centering
    \begin{tikzpicture}[
    align=center,font=\scriptsize,level distance=3.2em, 
    every node/.style={rounded corners=0.6ex},
    edge from parent/.style={draw=darkgrey,ultra thin},
    level 1/.style={sibling distance=7em}]
    % \coordinate
    \node(retriever)[fill=lightblue]{Retriever}
        child[edge from parent fork down]{node[fill=lightgrey]{Sparse\\Retriever}}
        child[edge from parent fork down]{node[fill=lightgrey]{Dense\\Retriever}}
        child[edge from parent fork down]{node[fill=lightgrey]{Iterative\\Retriever}};
    \node[fill=lightblue] (reader) [right=of retriever,xshift=11em]{Reader}
        child[edge from parent fork down]{node[fill=lightgrey]{Extractive\\Reader}}
        child[edge from parent fork down]{node[fill=lightgrey]{Generative\\Reader}};
        % [edge from parent fork down]
    \end{tikzpicture}
    \caption{The category of retriever and reader.}
    \label{fig:retriever-reader}
\end{figure}
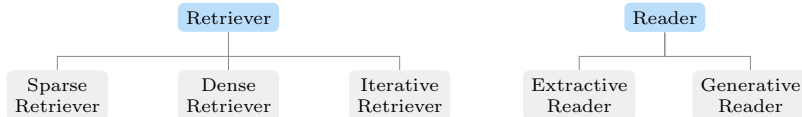

The methods discussed in the above subsections are used for closed-domain table QA. In this subsection, we discuss the retriever-reader architecture that is usually adopted for \textbf{open-domain table QA}, which provides answers by retrieval and reading. The retrieval model is in charge of retrieving the related documents containing tables from a large corpus, and the reader is used to produce the answers from the retrieved table documents. The retriever can be further categorized into \textit{sparse retriever}, \textit{dense retriever}, and \textit{iterative retriever} . The sparse/dense retriever uses sparse/dense representations for the question and candidate documents and then performs matching between them. The iterative retriever tries to retrieve the relevant documents in multiple steps, which appends the reliable retrieval from the previous step for the next step of retrieval. As for the reader, it can be further classified into \textit{generative reader} and \textit{extractive reader}, which adopt the generative method and extractive method discussed in subsections \ref{subsection:generative methods} and \ref{subsection:extractive methods}, respectively. We depict an overview of these categories in Figure \ref{fig:retriever-reader} and present the representative works below.

\textbf{Extractive Reader}: Chen et al. \cite{chenOpenQuestionAnswering2021} propose the fusion retriever (a dense retriever) and cross-block reader for OTT-QA task. The fusion retriever first pre-aligns the table segments to their related passages by entity linking and groups them into a \textit{fused block}. Then, it retrieves the top-$k$ fused blocks for cross-block reading that uses the long range Transformer\cite{Ainslie2020ETCEL}. Moreover, Zhong  et al. \cite{zhongReasoningHybridChain2022} also train a chain extractor as the auxiliary model for RoBERTa-based dense retriever and Longformer-based \cite{Beltagy2020LongformerTL} extractive reader. The chain extractor extracts possible reasoning chains from the table; then the reasoning chains are flattened and appended to the retrieved tables as the input of the reader, which leads to better answer selection. Instead of using general pre-trained models, Herzig et al \cite{herzigOpenDomainQuestion2021} adopt a table-oriented pre-trained model, TAPAS \cite{herzigTaPasWeaklySupervised2020}, as both the dense retriever and extractive reader to achieve better performance. 

\textbf{Generative Reader}: Li et al. \cite{Li2021DualRO} adopt a sparse retriever using BM25 to retrieve tables or text, in which the tables are flattened into passages by concatenating the cell values of each row. After retrieving the top-$k$ candidate tables or passages, a generative dual reader-parser that is based on FiD model \cite{Izacard2021LeveragingPR} is used to generate the answer directly or produce an SQL query. The FiD model is a Seq2Seq transformer that takes the question and top-$k$ candidates as input and fuses them in decoder for answer or logical form generation (referred to as Fusion in Decoder, FiD). Whether the output will be a final answer or SQL query is decided by the FiD model automatically based on the input question and its context. UniK-QA \cite{oguzUniKQAUnifiedRepresentations2021} also adopts a FiD model as the reader but takes a BERT-based dense retriever~\cite{DBLP:journals/corr/abs-1905-07588}. Moreover, it adapts the model to multiple knowledge sources, including text, tables, and knowledge base, by flattening the (semi-)structured data. 

The open-domain table QA task is a new challenge that has appeared in recent years. Most works \cite{herzigTaPasWeaklySupervised2020,Li2021DualRO,oguzUniKQAUnifiedRepresentations2021} simply flatten the tables and adapt methods used in open-domain text-based QA for this task. However, this may lead to the loss of important information in the (semi-)structured data \cite{oguzUniKQAUnifiedRepresentations2021}. Hence,  the encoding methods of tables can be further explored in the open-domain setting.

\section{Challenges and Future Directions}
\label{section:challengs}
This paper presents an overview of existing datasets and five different methods for table QA. Some methods have revealed desirable performance on relatively easier benchmarks. However, there are still several key challenges for future work, especially in table QA for non-database tables. In this survey, we share our thoughts on some of the main challenges in table QA.

\textbf{Numerical representation for table QA}: Numerical values are the common content of table cells, especially for spreadsheet tables. Dedicated numerical representation might be a key factor for non-database table QA. For example, Zhu et al. \cite{Zhu2021TATQAAQ} report that about 55\% of errors in the RoBERTa-based model (called TAGOP) are caused by incorrect evidence extraction, most of them arising because of premature numerical representation. Hence, it is an interesting challenge to incorporate better numerical representations into table QA models. 

\textbf{Complex reasoning in non-database table QA}: For database tables, researchers have developed semantic parsing-based methods for complex reasoning. However, most existing methods for non-database table QA only support simple reasoning. For example, TAGOP \cite{Zhu2021TATQAAQ} only support one-step operation, FinQANet \cite{chenFinQADatasetNumerical2021a} supports nested operations but limited to four basic arithmetics. Future works include how to design a more general logical form that could support complex reasoning on most non-database table QA tasks. 
%  We wonder if there may be a more general logical form that could support complex reasoning on most non-database table QA tasks and what model design may lead to better performance. We believe it is worthy of further exploration.

\bibliographystyle{splncs04}
\bibliography{refs.bib}

\end{document}